\documentclass[11pt]{article}
\usepackage{acl2016}
\usepackage{times}
\usepackage{latexsym}
\usepackage{multicol}
\usepackage{amsmath}
\usepackage{graphicx}
\usepackage{url}

\aclfinalcopy 

\newcommand{\mycomment}[1]{}
\newcommand{\ignore}[1]{}

\DeclareMathOperator*{\logodds}{\sigma^{-1}}

\newcommand{\entail}{{\Rightarrow}}
\newcommand{\notentail}{{\,/\!\!\!\!{\Rightarrow}}}

\newcommand{\loent}{{\tilde{\Rightarrow}}}

\newcommand{\fie}{{\raisebox{1.8pt}{\makebox[0pt][l]{\hspace{0.2pt}${\scriptscriptstyle <}$}${\scriptscriptstyle \bigcirc}$}}}
\newcommand{\bie}{{\raisebox{1.8pt}{\makebox[0pt][l]{\hspace{1.3pt}${\scriptscriptstyle >}$}${\scriptscriptstyle \bigcirc}$}}}
\newcommand{\ip}{{\cdot}}

\title{ A Vector Space for Distributional Semantics for Entailment
  \Thanks{This work was partially supported by French ANR grant CIFRE N 1324/2014.}
}

\author{James Henderson \and Diana Nicoleta Popa \\
  Xerox Research Centre Europe \\ 
  {\tt james.henderson@xrce.xerox.com} \and {\tt diana.popa@xrce.xerox.com}}

\date{}

\begin{document}
\thispagestyle{myheadings}
\pagenumbering{gobble} 
\markright{\protect\parbox{\textwidth}{\footnotesize
To appear in Proc.\ 54th Annual Meeting of the Association Computational Linguistics (ACL 2016). 
\hfill} } 
\maketitle

\begin{abstract}
Distributional semantics creates vector-space representations that 
capture many forms of semantic similarity, but their relation to semantic
entailment has been less clear.  We propose a vector-space model which
provides a formal foundation for a distributional semantics of entailment.
Using a mean-field approximation,
we develop approximate inference procedures and entailment operators over
vectors of probabilities of features being known (versus unknown).  We use
this framework to reinterpret an existing distributional-semantic model
(Word2Vec)
as approximating an entailment-based model of the distributions of words in
contexts, thereby predicting lexical entailment relations.  In both
unsupervised and semi-supervised experiments on hyponymy detection, we get
substantial improvements over previous results.

\end{abstract}

\section{Introduction}
\label{sec:intro}

Modelling entailment is a fundamental issue in computational semantics.  It is
also important for many applications, for example to produce abstract
summaries or to answer questions from text, where we need to ensure that the
input text entails the output text.  There has been a lot of interest in
modelling entailment in a vector-space, but most of this work takes an
empirical, often ad-hoc, approach to this problem, and achieving good results
has been difficult \cite{Levy2015}.  In this work, we propose a new framework
for modelling entailment in a vector-space, and illustrate its effectiveness
with a distributional-semantic model of hyponymy detection.

\begin{table}[tb]
\begin{center}
\begin{tabular}{r|c|c|c|c|}
\(\entail\) & {\em unk} & \(f\) & \(g\) & \(\neg f\) \\
\hline
{\em unk}	 & 1 & 0 & 0 & 0 \\
\hline
\(f\)	 & 1 & 1 & 0 & 0 \\
\hline
\(g\)	 & 1 & 0 & 1 & 0 \\
\hline
\(\neg f\)	 & 1 & 0 & 0 & 1 \\
\hline
\end{tabular}
\vspace*{-1ex}
\end{center}
\caption{ 
Pattern of logical entailment between nothing known ({\em unk}), two different features $f$ and $g$ known, and the complement of $f$ ($\neg f$) known.
\vspace{-1ex}
}
\label{tab:example}
\end{table}

Unlike previous vector-space models of entailment, the proposed framework
explicitly models what information is unknown.  This is a crucial property,
because entailment reflects what information is and is not known; a
representation $y$ entails 
a representation $x$ if and only if everything that is known given $x$ is also
known given $y$.  Thus, we model entailment in a vector space where each
dimension represents something we might know.  As
illustrated in Table~\ref{tab:example}, knowing that a feature $f$ is true
always entails knowing that same feature, but never entails knowing that a
different feature $g$ is true.  Also, knowing that a feature is true always
entails not knowing anything ({\em unk}), since strictly less information
is still entailment, but the reverse is never true.  Table~\ref{tab:example}
also illustrates that knowing that a feature $f$ is false ($\neg f$) patterns
exactly the same way as knowing that an unrelated feature $g$ is true.  This
illustrates that the relevant dichotomy for entailment is known versus
unknown, and not true versus false.

Previous vector-space models have been very successful at modelling semantic
similarity, in particular using distributional semantic models
(e.g.\ \cite{Deerwester1990,Schutze93,word2vec1}).  Distributional semantics
uses the distributions of words in contexts to induce vector-space embeddings
of words, which have been shown to be useful for a wide variety of tasks.  Two
words are predicted to be similar if the dot product between their vectors is
high.  But the dot product is an anti-symmetric operator, which makes it more
natural to interpret these vectors as representing whether features are true
or false, whereas the dichotomy known versus unknown is asymmetric.  We
surmise that this is why distributional semantic models have had difficulty
modelling lexical entailment \cite{Levy2015}.

To develop a vector-space model of whether features are known or unknown, we
start with discrete binary vectors, where 1 means known and 0 means unknown.
Entailment
between these discrete binary vectors can be calculated by independently
checking each dimension.
But as soon as we
try to do calculations with {\em distributions} over these vectors, we need to
deal with the case where the features are not independent.  For example, if
feature $f$ has a 50\% chance of being true and a 50\% chance of being false,
we can't assume that there is a 25\% chance that both $f$ and $\neg f$ are
known.
This simple case of mutual exclusion is just one example of a wide range of
constraints between features which we need to handle in semantic models.
These constraints mean that the different dimensions of our vector space are
not independent, and therefore exact models are not factorised.  Because the
models are not factorised, exact calculations of entailment and exact
inference of vectors are intractable.

Mean-field approximations are a popular approach to efficient inference for
intractable models.  In a mean-field approximation, distributions over binary
vectors are represented using a single probability for each dimension.
These vectors of real values are the basis of our proposed vector space for
entailment.

In this work, we propose a vector-space model which provides a formal
foundation for a distributional semantics of entailment.  This framework is
derived from a mean-field approximation to entailment between binary vectors,
and includes operators for measuring entailment between vectors, and
procedures for inferring vectors in an entailment graph.  We validate this
framework by using it to reinterpret existing Word2Vec \cite{word2vec1} word
embedding vectors as approximating an entailment-based model of the
distribution of words in contexts.  This reinterpretation allows us to use
existing word embeddings as an unsupervised model of lexical entailment,
successfully predicting hyponymy relations using the proposed entailment
operators in both unsupervised and semi-supervised experiments.

\section{Modelling Entailment in\,a\,Vector Space}
\label{sec:modelling}

To develop a model of entailment in a vector space, we start with the logical
definition of entailment in terms of vectors of discrete known features: $y$
entails $x$
if and only if all the known features in $x$ are also
included in $y$.  We formalise this relation with binary vectors $x,y$ where $1$
means known and $0$ means unknown, 
so this discrete entailment relation $(y\entail x)$ can be defined with the
binary formula:
\vspace{-1ex}
\[
P((y\entail x) ~|~ x,y) = \prod_k (1 - (1{-}y_k) x_k)
\vspace{-1ex}
\]
Given prior probability distributions $P(x),P(y)$ over these vectors,
the exact joint and marginal probabilities for an entailment relation are:
\vspace{-1ex}
\[
P(x, y, (y\entail x)) = P(x) ~P(y) \prod_k (1 {-} (1{-}y_k) x_k)
\vspace{-2.5ex}
\]
\begin{align}
\label{eqn:pent}
\!\,P((y\entail x)) = E_{P(x)} E_{P(y)} \prod_k (1 {-} (1{-}y_k) x_k)\!\!
\end{align}
\vspace{-2ex}

We cannot assume that the priors $P(x)$ and $P(y)$ are factorised, because
there are many important correlations between features and therefore we cannot
assume that the features are independent.  As discussed in
Section~\ref{sec:intro}, even just representing both a feature $f$ and its
negation $\neg f$ requires two different dimensions $k$ and $k^\prime$ in the
vector space, because $0$ represents unknown and not false.  Given valid
feature vectors, calculating entailment can consider these two dimensions
separately, but to reason with distributions over vectors we need the prior
$P(x)$ to enforce the constraint that $x_k$ and $x_{k^\prime}$ are mutually
exclusive.  In general, such correlations and anti-correlations exist between
many semantic features, which makes inference and calculating the probability
of entailment intractable.

To allow for efficient inference in such a model, we propose a mean-field
approximation.  This in effect assumes that the posterior distribution over
vectors is factorised, but in practice this is a much weaker assumption than
assuming the prior is factorised.  The posterior distribution has less
uncertainty and therefore is influenced less by non-factorised prior
constraints.  By assuming a factorised posterior, we can then represent
distributions over feature vectors with simple vectors of probabilities of
individual features (or as below, with their log-odds).  These real-valued
vectors are the basis of the proposed vector-space model of entailment.

In the next two subsections, we derive a mean-field approximation for
inference of real-valued vectors in entailment graphs.  This derivation leads
to three proposed vector-space operators for approximating the log-probability
of entailment, summarised in Table~\ref{tab:ops}.  These operators will be
used in the evaluation in Section~\ref{sec:evaluation}.  This inference
framework will also be used in Section~\ref{sec:interpreting} to model how
existing word embeddings can be mapped to vectors to which the entailment
operators can be applied.  

\subsection{A Mean-Field Approximation}
\label{sec:mean-filed}

A mean-field approximation approximates the posterior $P$ using a factorised
distribution $Q$.  First of all, this gives us a concise description of the
posterior $P(x|\ldots)$ as a vector of continuous values $Q(x{=}1)$, where
$Q(x{=}1)_k=Q(x_k{=}1)\approx E_{P(x|\ldots)}x_k=P(x_k{=}1|\ldots)$ (i.e.\ the
marginal probabilities of each bit).  Secondly, as is shown below, this gives
us efficient methods for doing approximate inference of vectors in a model.

First we consider the simple case where we want to approximate the posterior
distribution $P(x,y|y\entail x)$.  In a mean-field approximation, we want to
find a factorised distribution $Q(x,y)$ which minimises the KL-divergence
$D_{KL}(Q(x,y)||P(x,y|y\entail x))$ with the true distribution $P(x,y|y\entail
x)$.
\vspace{-1ex}
\begin{align*}
  L ~~&\!\!=~ D_{KL}(Q(x,y)||P(x,y|(y\entail x))) 
  \\
  &\hspace{-1.5ex}
  \propto \sum_x Q(x)\log\frac{Q(x,y)}{P(x,y,(y\entail x))}
    \\
    =& 
      \sum_k E_{Q(x_{k})}\log Q(x_{k}) + \sum_k E_{Q(y_{k})}\log Q(y_{k}) 
      \\&
      - E_{Q(x)}\log P(x) - E_{Q(y)}\log P(y)
      \\&
      - \sum_k E_{Q(x_k)}E_{Q(y_k)}\log(1{-}(1{-}y_{k})x_{k}) 
\end{align*}
\vspace{-3ex}\newline
In the final equation, the first two terms are the negative entropy of $Q$,
$-H(Q)$, which acts as a maximum entropy regulariser, the final term enforces
the entailment constraint, and the middle two terms represent the prior for
$x$ and $y$.  One approach (generalised further in the next subsection) to
the prior terms ${-}E_{Q(x)}\log P(x)$ is to bound them by assuming $P(x)$ is
a function in the exponential family,
giving us:
\vspace{-1ex}
\begin{align*}
  E_{Q(x)}\log P(x) 
  &\geq E_{Q(x)}\log \frac{\exp( \sum_k \theta^x_k x_k )}{{\cal Z}_\theta}
    \\
    &=~ \sum_k E_{Q(x_k)}\theta^x_k x_k - \log {\cal Z}_\theta
\end{align*}
\vspace{-2.5ex}\newline
where the $\log {\cal Z}_\theta$ is not relevant in any of our inference
problems and thus will be dropped below.

As typically in mean-field approximations, inference of $Q(x)$ and $Q(y)$
can't be done efficiently with this exact objective $L$, because of the
nonlinear interdependence between $x_k$ and $y_k$ in the last term.  Thus, we
introduce two approximations to $L$, one for use in inferring $Q(x)$ given
$Q(y)$ (forward inference), and one for the reverse inference problem
(backward inference).  In both cases, the approximation is done with an
application of Jensen's inequality to the log function, which gives us an
upper bound on $L$, as is standard practice in mean-field approximations.
For forward inference:
\vspace{-1ex}
\begin{align}
\label{eqn:fieL}
  L 
  \leq&
  -H(Q)
      - Q(x_k{=}1)\theta^x_k - E_{Q(y_k)}\theta^y_k y_k
      \\& \nonumber
      - Q(x_k{=}1)\log Q(y_k{=}1) ~)
\end{align}
\vspace{-3.5ex}\newline
which we can optimise for $Q(x_k{=}1)$:
\vspace{-1ex}
\begin{align}
  \label{eqn:forward}
  Q(x_k{=}1) 
  =
    \sigma(~ \theta^x_k + \log Q(y_k{=}1) ~)
\end{align}
\vspace{-3.5ex}\newline
where $\sigma()$ is the sigmoid function.  The sigmoid function arises from
the entropy regulariser, making this a specific form of maximum entropy model.
And for backward inference:
\vspace{-1ex}
\begin{align}
\label{eqn:bieL}
  L 
  \leq& 
  -H(Q)
      - E_{Q(x_k)}\theta^x_k x_k - Q(y_k{=}1)\theta^y_k
      \\& \nonumber
      - (1{-}Q(y_k{=}1))\log (1{-}Q(x_k{=}1)) ~)
\end{align}
\vspace{-3.5ex}\newline
which we can optimise for $Q(y_k{=}1)$:
\vspace{-1ex}
\begin{align}
  \label{eqn:backward}
  Q(y_k{=}1) 
  =~
    \sigma(~ \theta^y_k - \log (1{-}Q(x_k{=}1)) ~)
\end{align}
\vspace{-3ex}

Note that in equations (\ref{eqn:fieL}) and (\ref{eqn:bieL}) the final terms,
$Q(x_k{=}1)\log Q(y_k{=}1)$ and \linebreak
$(1{-}Q(y_k{=}1))\log (1{-}Q(x_k{=}1))$
respectively, are approximations to the log-probability of the entailment.  We
define two vector-space operators, $\fie$ and $\bie$, to be these same
approximations.
\vspace{-1ex}
\begin{align*}
  \log Q(y\entail x) 
  \hspace*{-1.3cm} &
  \\
  &\approx \sum_k
    E_{Q(x_k)} \log( E_{Q(y_k)} (1 - (1{-}y_k) x_k) )
  \\
  &=~  Q(x{=}1) \cdot \log Q(y{=}1)
  ~\equiv
  X \fie Y 
\\[1ex]
  \log Q(y\entail x) 
  \hspace*{-1.3cm} &
  \\
  &\approx \sum_k
    E_{Q(y_k)} \log( E_{Q(x_k)} (1 - (1{-}y_k) x_k) )
  \\
  &=~  (1{-}Q(y{=}1)) \cdot \log (1{-}Q(x{=}1))
  ~\equiv
  Y \bie X 
\end{align*}
\vspace{-3ex}

We parametrise these
operators with the vectors $X,Y$ of log-odds of $Q(x),Q(y)$, namely
$
X = \log\frac{Q(x{=}1)}{Q(x{=}0)} = \logodds(Q(x{=}1))
$.
\vspace{0.5ex}
The resulting operator definitions are summarised in Table~\ref{tab:ops}.

\begin{table}[tb]
\begin{center}
\fbox{
\vspace{-3ex}
\parbox[b][12ex]{0.8\columnwidth}{
\begin{align*}
  X \fie Y 
  ~\equiv~~& \sigma(X) \cdot \log \sigma(Y) 
\\
  Y \bie X 
  ~\equiv~~&  \sigma(-Y) \cdot \log \sigma(-X) 
\\
  Y \loent X 
  ~\equiv~~&  \sum_k \log( 1 - \sigma(-Y_k) \sigma(X_k) ) 
\\[-6.2ex]
\end{align*}
} }
\vspace{-4ex}\newline
\end{center}
\caption{ 
The proposed entailment operators, approximating $\log P(y\entail x)$.
\vspace{-1ex}
}
\label{tab:ops}
\end{table}

Also note that the probability of entailment given in equation
(\ref{eqn:pent}) becomes factorised when we replace $P$ with $Q$.  We define a
third vector-space operator, $\loent$, to be this factorised approximation,
also shown in Table~\ref{tab:ops}.

\subsection{Inference in Entailment Graphs}
\label{sec:inference}

In general, doing inference for one entailment is not enough; we
want to do inference in a graph of entailments between variables.  In this
section we generalise the above
mean-field approximation to
entailment graphs.

To represent information about variables that comes from outside the
entailment graph, we assume we are given a prior $P(x)$ over all variables
$x_i$ in the graph.  As above, we do not assume that this prior is factorised.
Instead we assume that the prior $P(x)$ is itself a graphical model which can
be approximated with a mean-field approximation.  

Given a set of variables $x_i$ each representing vectors of binary variables
$x_{ik}$, a set of entailment relations $r=\{(i,j)|(x_i\entail x_j)\}$, and a
set of negated entailment relations $\bar{r}=\{(i,j)|(x_i\notentail x_j)\}$, we
can write the joint posterior probability as:
\vspace{-1ex}
\begin{align*}
  P(x,r,\bar{r}) 
  = ~\frac{1}{\cal Z} P(x) \prod_i \Big(~
  \hspace*{-2.9cm} &
  \\&
  ( \prod_{j:r(i,j)} \prod_k P(x_{ik}\entail x_{jk}|x_{ik},x_{jk}) )
  \\&
  ( \prod_{j:\bar{r}(i,j)} (1 - \prod_k P(x_{ik}\entail x_{jk}|x_{ik},x_{jk})) )
  ~\Big)
\end{align*}
\vspace{-3ex}\newline
We want to find a factorised distribution $Q$ that minimises
$L = D_{KL}(Q(x)||P(x|r,\bar{r}))$.  
As above, we bound this loss for each element $X_{ik}{=}\logodds(Q(x_{ik}{=}1))$ of
each vector we want to infer, using analogous Jensen's inequalities for the
terms involving nodes $i$ and $j$ such that $r(i,j)$ or $r(j,i)$.  For
completeness, we also propose similar inequalities for nodes $i$ and $j$ such
that $\bar{r}(i,j)$ or $\bar{r}(j,i)$, and bound them using the constants
\vspace{-1ex}
\[
C_{ijk} \geq \prod_{k^\prime\neq k}
(1{-}\sigma(-X_{ik^\prime})\sigma(X_{jk^\prime}))
.
\vspace{-1ex}
\]

To represent the prior $P(x)$, we use the terms 
\vspace{-1ex}
\[
\theta_{ik}(X_{\bar{ik}}) \leq \log
\frac{E_{Q(x_{\bar{ik}})} P(x_{\bar{ik}},x_{ik}{=}1)}
     { 1 - E_{Q(x_{\bar{ik}})} P(x_{\bar{ik}},x_{ik}{=}1)} 
\vspace{-1ex}
\]
where $x_{\bar{ik}}$ is the set of all $x_{i^\prime k^\prime}$ such that
either $i^\prime{\neq}i$ or $k^\prime{\neq}k$.  These terms can be thought of
as the log-odds terms that would be contributed to the loss function by
including the prior's graphical model in the mean-field approximation.

Now we can infer the optimal $X_{ik}$ as:
\vspace{-1ex}
\begin{align}
  &X_{ik} 
  ~=~
  \theta_{ik}(X_{\bar{ik}})
  ~+ \!\sum_{j:r(i,j)}\! {-}\log\sigma({-}X_{jk})
  \\[-0.5ex]&\nonumber
  ~~+ \!\!\sum_{j:r(j,i)} \log\sigma(X_{jk})
  ~+ \!\!\sum_{j:\bar{r}(j,i)} \log\frac{1 {-} C_{ijk} \sigma(X_{jk})}{1 {-} C_{ijk}}
  \\&\nonumber
  ~~+ \!\sum_{j:\bar{r}(i,j)}\! {-}\log\frac{1 {-} C_{ijk} \sigma({-}X_{jk})}{1 {-} C_{ijk}}
\label{eqn:graph-inference}
\end{align}
\vspace{-2ex}

In summary, the proposed mean-field approximation does inference in entailment
graphs by iteratively re-estimating each $X_i$ as 
the sum of: the prior log-odds, ${-}\log\sigma(-X_{j})$ for each entailed
variable $j$, and $\log\sigma(X_{j})$ for each entailing variable
$j$.\footnote{It is interesting to note that ${-}\log\sigma(-X_{j})$ is a
  non-negative transform of $X_j$, similar to the ReLU nonlinearity which is
  popular in deep neural networks \cite{Glorot11}.
  $\log\sigma(X_{j})$ is the analogous non-positive transform.}  This
inference optimises $X_i\fie X_j$ for each entailing $j$ plus $X_i\bie X_j$
for each entailed $j$, plus a maximum entropy regulariser on $X_i$.  Negative
entailment relations, if they exist, can also be incorporated with some
additional approximations.  Complex priors can also be incorporated through
their log-odds, simulating the inclusion of the prior within the mean-field
approximation.

Given its dependence on mean-field approximations, it is an empirical question
to what extent we should view this model as computing real entailment
probabilities and to what extent we should view it as a well-motivated
non-linear mapping for which we simply optimise the input-output behaviour (as
for neural networks \cite{Henderson10_jmlr}).  In
Sections~\ref{sec:interpreting} and~\ref{sec:evaluation} we argue for the
former (stronger) view.

\section{Interpreting Word2Vec Vectors}
\label{sec:interpreting}

To evaluate how well the proposed framework provides a formal foundation for
the distributional semantics of entailment, we use it to re-interpret an
existing model of distributional semantics in terms of semantic entailment.
There has been a lot of work on how to use the distribution of
contexts in which a word occurs to induce a vector representation of
the semantics of words.  In this paper, we leverage this previous work
on distributional semantics by re-interpreting a previous
distributional semantic model and using this understanding to map its
vector-space word embeddings to vectors in the proposed framework.  We
then use the proposed operators to predict entailment between words
using these vectors.  In Section~\ref{sec:evaluation} below, we
evaluate these predictions on the task of hyponymy detection.
In this section we motivate three different ways to interpret the Word2Vec
\cite{word2vec1,word2vec2_nips} distributional semantic model as an
approximation to an entailment-based model of the semantic relationship
between a word and its context.

Distributional semantics learns the semantics of words by looking at the
distribution of contexts in which they occur.  To model this relationship, we
assume that the semantic features of a word are (statistically speaking)
redundant with those of its context words, and consistent with those of its
context words.  We model these properties using a hidden vector which is the
consistent unification of the features of the middle word and the context.  In
other words, there must exist a hidden vector which entails both of these
vectors, and is consistent with prior constraints on vectors.  We split this
into two steps, inference of the hidden vector $Y$ from the middle vector
$X_m$, context vectors $X_c$ and prior, and computing the log-probability
(\ref{eqn:distsem}) that this hidden vector entails the middle and context
vectors:
\vspace{-0.5ex}
\begin{equation}
\label{eqn:distsem}
\max_Y( \log P(y,y\entail x_m,y\entail x_c) ) 
\end{equation}

We interpret Word2Vec's Skip-Gram model as learning its context and middle
word vectors so that the log-probability of this entailment is high for the
observed context words and low for other (sampled) context words.  The word
embeddings produced by Word2Vec are only related to the vectors $X_m$ assigned
to the middle words; context vectors are computed but not output.
We model the context vectors $X_c^\prime$ 
as combining (as in equation~(\ref{eqn:backward})) information about a context
word itself with information which can be inferred from this word given the
prior, $ X_c^\prime = \theta_c - \log\sigma(-X_c) $.

The numbers in the vectors output by Word2Vec are real numbers between negative
infinity and infinity, so the simplest interpretation of them is as the
log-odds of a feature being known.  In this case we can treat these vectors
directly as the $X_m$ in the model.  
The inferred hidden vector $Y$ can then be calculated
using the model of backward inference from the previous section.
\vspace{-1ex}
\begin{align*}
Y &= \theta_c - \log\sigma(-X_c) - \log\sigma(-X_m)
\\&
= X_c^\prime  - \log\sigma(-X_m)
\end{align*}
\vspace{-3ex}\newline
Since the unification $Y$ of context and middle word features is computed
using backward inference, we use the backward-inference operator $\bie$ to
calculate how successful that unification was.  This gives us the final score:
\vspace{-1ex}
\begin{align*}
\log P(y,y\entail x_m,y\entail x_c) 
\hspace*{-2.5cm} &
\\\approx~&
Y\bie X_m + Y\bie X_c + {-}\sigma({-}Y)\ip \theta_c
\\=~&
Y\bie X_m + {-}\sigma({-}Y)\ip X_c^\prime
\end{align*}
\vspace{-3ex}

This is a natural interpretation, but it ignores the equivalence in Word2Vec
between pairs of positive values and pairs of negative values, due to its use
of the dot product.  As a more accurate interpretation, we interpret each
Word2Vec dimension as specifying whether its feature is known to be true or
known to be false.  Translating this Word2Vec vector into a vector in our entailment vector space,
we get one copy $Y^+$ of the vector representing known-to-be-true features and a
second negated duplicate $Y^-$ of the vector representing known-to-be-false
features, which we concatenate to get our representation $Y$.
\vspace{-1ex}
\begin{align*}
Y^+ =~& X_c^\prime  - \log\sigma(-X_m)
\\
Y^- =~& -X_c^\prime  - \log\sigma(X_m)
\\
\log P(y,y\entail x_m,y\entail x_c) 
\hspace*{-2.6cm} &
\\ \approx~&
Y^+\bie X_m + {-}\sigma({-}Y^+)\ip X_c^\prime
\\&
+ Y^-\bie (-X_m) + {-}\sigma({-}Y^-)\ip (-X_c^\prime)
\end{align*}
\vspace{-3ex}

As a third alternative, we modify this latter interpretation with some
probability mass reserved for unknown in the vicinity of zero.  By subtracting
1 from both the original and negated copies of each dimension, we get a
probability of unknown of $1{-}\sigma(X_m{-}1)-\sigma(-X_m{-}1)$.
This gives us:
\vspace{-1ex}
\begin{align*}
Y^+ =~& X_c^\prime  - \log\sigma(-(X_m{-}1))
\\
Y^- =~& -X_c^\prime  - \log\sigma(-(-X_m{-}1))
\\
\log P(y,y\entail x_m,y\entail x_c) 
\hspace*{-2.6cm} &
\\ \approx~&
Y^+\bie (X_m{-}1) + {-}\sigma({-}Y^+)\ip X_c^\prime
\\&
+ Y^-\bie (-X_m{-}1)) + {-}\sigma({-}Y^-)\ip (-X_c^\prime)
\end{align*}
\vspace{-3ex}

\begin{figure}[tb]
  \begin{center}
    \includegraphics[width=1.\linewidth]{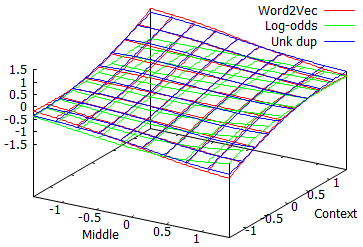}
    \vspace*{-5ex}
  \end{center}
  \caption{  The learning gradients for Word2Vec, the {\em log-odds} $\bie$, and the {\em unk dup} $\bie$
    interpretation of its vectors.  
}
\label{fig:gradients}
\end{figure}

To understand better the relative accuracy of these three interpretations, we
compared the training gradient which Word2Vec uses to train its middle-word
vectors to the training gradient for each of these interpretations.  We
plotted these gradients for the range of values typically found in Word2Vec
vectors for both the middle vector and the context
vector.  Figure~\ref{fig:gradients} shows three of these plots.
  As expected, the second
interpretation is more accurate than the 
first because its plot is anti-symmetric around the diagonal, like the
Word2Vec gradient.  In the third alternative, the constant 1 was chosen to
optimise this match, 
producing a close match to the Word2Vec training gradient, as shown in
Figure~\ref{fig:gradients} ({\em Word2Vec} versus {\em Unk dup}).

Thus, Word2Vec can be seen as a good
approximation to the third model, and a progressively worse approximation to
the second and first models.  Therefore, if the entailment-based
distributional semantic model we propose is accurate, then we would expect the
best accuracy in hyponymy detection using the third interpretation of
Word2Vec vectors, and progressively worse accuracy for the other two interpretations.
As we will see in Section~\ref{sec:evaluation}, this prediction holds.

\section{Related Work}
\label{sec:related}


There has been a significant amount of work on using distributional-semantic
vectors for hyponymy detection, using supervised, semi-supervised or
unsupervised methods
(e.g.\ \cite{ZhengICAJ15,Necsulescu_15,Vylomova_15,weeds2014learning,FuGuo,rei2014looking}).
Because our main concern is modelling entailment within a vector space, we do
not do a thorough comparison to models which use measures computed outside the vector space
(e.g.\ symmetric measures (LIN \cite{Lin1998}), asymmetric measures
(WeedsPrec \cite{Weeds2003,Weeds2004}, balAPinc \cite{Kotlerman2010}, invCL \cite{Lenci2012}) and entropy-based measures (SLQS \cite{Santus14})),
nor to models which encode hyponymy in the parameters of a vector-space
operator or classifier \cite{FuGuo,roller2014,Baroni12}).  
We also limit our evaluation of lexical entailment to hyponymy, not including
other related lexical relations (cf.\ 
\cite{weeds2014learning,Vylomova_15,Turney2014,Levy14_conll}), leaving more
complex cases to future work on compositional semantics.
We are also not concerned with models or evaluations which require
supervised learning about individual words, instead limiting ourselves to
semi-supervised learning where the words in the training and test sets are
disjoint.

For these reasons, in our evaluations we replicate the experimental setup of
\newcite{weeds2014learning}, for both unsupervised and semi-supervised
models.  Within this setup, we compare to the results of the models evaluated by
\newcite{weeds2014learning} and to 
previously proposed vector-space operators.
This includes one vector space operator for hyponymy which doesn't have trained
parameters, proposed by \newcite{rei2014looking}, called {\em weighted cosine}.
The dimensions of the dot product (normalised to make it a cosine measure) are
weighted to put more weight on the larger values in the entailed (hypernym)
vector.  

We base this evaluation on the Word2Vec \cite{word2vec1,word2vec2_nips}
distributional semantic model and its publicly available word
embeddings.  
We choose it because it is popular, simple, fast, and its
embeddings have been derived from a very large corpus.  \newcite{Levy2014NIPS}
showed that it is closely related to the previous PMI-based
distributional semantic models (e.g.\ \cite{Turney2010}).

The most similar previous work, in terms of motivation and aims, is that of
\newcite{Vilnis15}.  They also model entailment directly using a vector space,
without training a classifier.  But instead of representing words as a point
in a vector space (as in this work), they represent words as a Gaussian
distribution over points in a vector space.  This allows them to represent the
extent to which a feature is known versus unknown as the amount of variance in
the distribution for that feature's dimension.  While nicely motivated
theoretically, the model appears to be more computationally expensive than the
one proposed here, particularly for inferring vectors.  They do make
unsupervised predictions of hyponymy relations with their learned vector
distributions, using KL-divergence between the distributions for the two
words.  They evaluate their models on the hyponymy data from \cite{Baroni12}.
As discussed further in section~\ref{sec:unsuper}, our best models achieve
non-significantly better average precision than their best models.

The semi-supervised model of \newcite{Kruszewski15} also models entailment in
a vector space, but they use a discrete vector space.  They train a mapping
from distributional semantic vectors to Boolean vectors such that feature
inclusion respects a training set of entailment relations.  They then use
feature inclusion to predict hyponymy, and other lexical entailment relations.
This approach is similar to the one used in our semi-supervised
experiments, except that their discrete entailment prediction operator is 
very different from our proposed entailment operators.  

\section{Evaluation}
\label{sec:evaluation}

To evaluate whether the proposed framework is an effective model of entailment
in vector spaces, we apply the interpretations from
Section~\ref{sec:interpreting} to publicly available word embeddings and use
them to predict the hyponymy relations in a benchmark dataset.  This framework
predicts that the more accurate interpretations of Word2Vec result in more
accurate unsupervised models of hyponymy.  We evaluate on detecting hyponymy
relations between words because hyponymy is the canonical type of lexical
entailment;
most of the semantic features of a hypernym (e.g.\ ``animal'') must be
included in the semantic features of the hyponym (e.g.\ ``dog'').  We evaluate
in both a fully unsupervised setup and a semi-supervised setup.


\subsection{Hyponymy with Word2Vec Vectors} 

For our evaluation on hyponymy detection, we replicate the experimental setup
of \newcite{weeds2014learning}, using their selection of word
pairs\footnote{\url{https://github.com/SussexCompSem/learninghypernyms}} from
the BLESS dataset \cite{bless-dataset}.\footnote{Of the 1667 word pairs in
  this data, 24 were removed because we do not have an embedding for one of
  the words.}  These noun-noun word pairs include positive hyponymy pairs,
plus negative pairs consisting of some other hyponymy pairs reversed, some
pairs in other semantic relations, and some random pairs.  Their selection is
balanced between positive and negative examples, so that accuracy can be used
as the performance measure.
For their semi-supervised experiments, ten-fold cross validation is
used, where for each test set, items are removed from the associated
training set if they contain any word from the test set.  Thus, the
vocabulary of the training and testing sets are always disjoint, thereby
requiring that the models learn about the vector space and not about the words
themselves.  We had to perform our own 10-fold split, but apply the same procedure to
filter the training set.

We could not replicate the word embeddings used in \newcite{weeds2014learning},
so instead we use publicly available word
embeddings.\footnote{\url{https://code.google.com/archive/p/word2vec/}}  These
vectors were trained with the Word2Vec software applied to about 100 billion
words of the Google-News dataset, and have 300 dimensions. 

The hyponymy detection results are given in Table~\ref{tab:results}, including
both unsupervised (upper box) and semi-supervised (lower box) experiments.  We
report two measures of performance, hyponymy detection accuracy ({\em 50\%
  Acc}\/) and direction classification accuracy ({\em Dir Acc}).  Since all
the operators only determine a score, we need to choose a threshold to get
detection accuracies.  Given that the proportion of positive examples in the
dataset has been artificially set at 50\%, we threshold each model's score
at the point where the proportion of positive examples output is 50\%, which
we call ``{\em 50\% Acc}\/''.  Thus the threshold is set after seeing the
testing inputs but not their target labels. 

 

\begin{table}[tb]
\begin{center}
\begin{tabular}{|@{~}c@{}c@{~}|ll@{~}|}
\hline
operator & supervision & \!\!50\% Acc & \!\!Dir Acc\\
\hline\hline
Weeds et.al. & None & 	58\% & 	~-- \\
\hline
{\em log-odds} $\fie$ & None & 	54.0\% & 	55.9\% \\
{\em weighted cos} & None & 	55.5\% & 	57.9\% \\
{\em dot} & None & 	56.3\% &	50\% \\
{\em dif} & None & 	56.9\% & 	59.6\% \\
{\em log-odds} $\loent$ & None & 	57.0\% & 	59.4\% \\
{\em log-odds} $\bie$ & None & 	60.1\%* & 	62.2\% \\
{\em dup} $\bie$ & None & 	61.7\% & 	68.8\% \\
{\em unk dup} $\loent$ & None & 	63.4\%* & 	68.8\% \\
{\em unk dup} $\bie$ & None & 	64.5\% & 	68.8\% \\
\hline\hline
Weeds et.al.           & SVM  & 	75\% & 	~-- \\
\hline
{\em mapped} {\em dif} & cross ent & 	64.3\% & 	72.3\% \\
{\em mapped} $\fie$  & cross ent & 	74.5\% & 	91.0\% \\
{\em mapped} $\loent$ & cross ent & 	77.5\% & 	92.3\% \\
{\em mapped} $\bie$  & cross ent & 	80.1\% & 	90.0\% \\
\hline
\end{tabular}
\end{center}
\caption{ Accuracies on the BLESS data from \newcite{weeds2014learning}, 
  for hyponymy detection ({\em 50\%  Acc}) and hyponymy direction
  classification ({\em Dir Acc}), in the unsupervised (upper box) and
  semi-supervised (lower box) experiments.  
  For unsupervised accuracies, * marks a significant difference with the 
  previous row. 
}
\label{tab:results}
\end{table}

Direction classification accuracy ({\em Dir Acc}) indicates how well the
method distinguishes the relative abstractness of two nouns.  Given a pair of
nouns which are in a hyponymy relation, it classifies which word is the
hypernym and which is the hyponym.  This measure only considers positive
examples and chooses one of two directions, so it is inherently a balanced
binary classification task.  Classification is performed by simply comparing
the scores in both directions.  If both directions produce the same score, the
expected random accuracy (50\%) is used.

As representative of previous work, we report the best results from
\newcite{weeds2014learning}, who try a number of unsupervised and semi-supervised
models, and use the same testing methodology and hyponymy data.  However, note
that their word embeddings are different.  For the semi-supervised models,
\newcite{weeds2014learning} trains classifiers, which are potentially more
powerful than our linear vector mappings.  We also compare the proposed operators to
the dot product ({\em dot}),\footnote{We also tested the cosine measure, but
  results were very slightly worse than {\em dot}.} vector differences ({\em dif}\/), and the
weighted cosine of \newcite{rei2014looking} ({\em weighted cos}), all computed
with the same word embeddings as for the proposed operators.

In Section~\ref{sec:interpreting} we argued for three progressively more
accurate interpretations of Word2Vec vectors in the proposed framework, the log-odds
interpretation ({\em log-odds} $\bie$),
the negated duplicate interpretation ({\em dup} $\bie$), and the negated
duplicate interpretation with unknown around zero ({\em unk dup} $\bie$).  We
also evaluate using the factorised calculation of entailment
({\em log-odds} $\loent$, 
{\em unk dup} $\loent$), and the backward-inference entailment operator 
({\em log-odds} $\fie$), neither of which match the proposed interpretations. 
For the semi-supervised case, we train a linear vector-space mapping 
into a new vector space, in which we apply the operators ({\em mapped} operators).  All
these results are discussed in the next two subsections.

\subsection{Unsupervised Hyponymy Detection} 
\label{sec:unsuper}

The first set of experiments evaluate the vector-space operators in
unsupervised models of hyponymy detection.
The proposed models are compared to
the dot product, because this is the standard vector-space operator and has
been shown to capture semantic similarity very well.  However, because the dot
product is a symmetric operator, it always performs at chance for direction
classification.  Another vector-space operator which has received much
attention recently is vector differences.  This is used (with vector sum) to
perform semantic transforms, such as ``king - male + female =
queen'', and has previously been used for modelling hyponymy
\cite{Vylomova_15,weeds2014learning}.  For our purposes, we sum the pairwise
differences to get a score which we use for hyponymy detection.

For the unsupervised results in the upper box of table~\ref{tab:results}, the
best unsupervised model of \newcite{weeds2014learning}, and the operators {\em
  dot}, {\em dif} and {\em weighted cos} all perform similarly on accuracy, as
does the 
log-odds 
factorised entailment calculation ({\em log-odds} $\loent$).  The
forward-inference entailment operator ({\em log-odds} $\fie$) performs above
chance but not well, as expected given the backward-inference-based
interpretation of Word2Vec vectors.  By definition, {\em dot} is at chance for
direction classification, but the other models all perform better, indicating
that all these operators are able to measure relative abstractness.  As
predicted, the $\bie$ operator performs significantly better than all these
results on accuracy, as well as on direction classification, even assuming the
log-odds interpretation of Word2Vec vectors.

When we move to the more accurate interpretation of Word2Vec vectors as
specifying both original and negated features ({\em dup} $\bie$), we improve
(non-significantly) on the log-odds interpretation.  Finally, the third and
most accurate interpretation, where values around zero can be unknown ({\em
  unk dup} $\bie$), achieves the best results in unsupervised hyponymy
detection, as well as for direction classification.  Changing to the
factorised entailment operator ({\em unk dup} $\loent$) is worse but also
significantly better than the other accuracies.

To allow a direct comparison to the model of \newcite{Vilnis15}, we also
evaluated the unsupervised models on the hyponymy data from \cite{Baroni12}.
Our best model achieved 81\% average precision on this dataset,
non-significantly better than the 80\% achieved by the best
model of \newcite{Vilnis15}.

\subsection{Semi-supervised Hyponymy Detection} 

Since the unsupervised learning of word embeddings may reflect many
context-word correlations which have nothing to do with hyponymy, we also
consider a semi-supervised setting.  Adding some supervision helps distinguish
features that capture semantic properties from other features which are not
relevant to hyponymy detection.  But even with supervision, we still want the
resulting model to be captured in a vector space, and not in a parametrised
scoring function.  Thus, we train mappings from the Word2Vec word vectors to
new word vectors, and then apply the entailment operators in this new vector
space to predict hyponymy.  Because the words in the testing set are always
disjoint from the words in the training set, this experiment measures how well
the original unsupervised vector space captures features that generalise
entailment across words, and not how well the mapping can learn about
individual words.  

Our objective is to learn a mapping to a new vector space in which an operator
can be applied to predict hyponymy.  We train linear mappings for the $\bie$
operator ({\em mapped} $\bie$) and for vector differences ({\em mapped} {\em
  dif}\/), since these were the best performing proposed operator and baseline
operator, respectively, in the unsupervised experiments.  We do not use the
duplicated interpretations 
because these transforms are
subsumed by the ability to learn a linear mapping.\footnote{Empirical results
  confirm that this is in practice the case, so we do not include these
  results in the table.}  Previous work on using vector differences for
semi-supervised hyponymy detection has used a linear SVM
\cite{Vylomova_15,weeds2014learning}, which is mathematically equivalent to
our vector-differences model, except that we use cross entropy loss and they
use a large-margin loss and SVM training.

The semi-supervised results in the bottom box of table~\ref{tab:results}
show a similar pattern to the unsupervised results.\footnote{It is not clear
  how to measure significance for cross-validation results, so we do not
  attempt to do so.}  The $\bie$ operator
achieves the best generalisation from training word vectors to testing word
vectors.  The {\em mapped} $\bie$ model has the best accuracy, followed by the
factorised entailment operator {\em mapped} $\loent$ and
\newcite{weeds2014learning}. 
Direction accuracies of all the proposed operators ({\em mapped} $\bie$, {\em
  mapped} $\loent$, {\em mapped} $\fie$) 
reach into the 90's.
The {\em dif} operator performs particularly poorly in this {\em mapped}
setting, perhaps because both the mapping and the operator are linear.
These semi-supervised results again support our distributional-semantic
interpretations of Word2Vec vectors and their associated entailment operator
$\bie$.

\section{Conclusion}

In this work, we propose a vector-space model which provides a formal
foundation for a distributional semantics of entailment.  
We developed a mean-field approximation to probabilistic
entailment between vectors which represent known versus unknown features.  And
we used this framework to derive vector operators for
entailment and vector inference equations for entailment graphs.   This
framework allows us to reinterpret Word2Vec as approximating an entailment-based distributional semantic model of
words in context,
and show that more accurate interpretations result in more accurate
unsupervised models of lexical entailment, achieving better accuracies than
previous models.  Semi-supervised evaluations confirm these results.

A crucial distinction between the semi-supervised models here and much previous work
is that they learn a mapping into a vector space which represents entailment,
rather than learning a parametrised entailment classifier.  Within this new
vector space, the entailment operators and inference equations apply, thereby
generalising naturally from these lexical representations to the compositional
semantics
of multi-word expressions and sentences.  Further work is needed to explore
the full power of these abilities to extract information about entailment from
both unlabelled text and labelled entailment data, encode it all in a single
vector space, and efficiently perform complex inferences about vectors and entailments.  This
future work on compositional distributional semantics should further
demonstrate the full power of the proposed framework for modelling
entailment in a vector space.


\newpage
\bibliography{entops_acl2016_final}
\bibliographystyle{acl2016}


\end{document}